\documentclass{article}

\usepackage{microtype}
\usepackage{graphicx}
\usepackage{subfigure}
\usepackage{booktabs} % for professional tables
\usepackage{amssymb}

\usepackage{hyperref}

\usepackage[accepted]{icml2021}

\icmltitlerunning{ThetA - fast and robust clustering via a distance parameter}

\begin{document}

\twocolumn[
\icmltitle{ThetA - fast and robust clustering via a distance parameter}



\icmlsetsymbol{equal}{*}

\begin{icmlauthorlist}
\icmlauthor{Eleftherios Garyfallidis}{iu}
\icmlauthor{Shreyas Fadnavis}{iu}
\icmlauthor{Jong Sung Park}{iu}
\icmlauthor{Bramsh Qamar Chandio}{iu}
\icmlauthor{Javier Guaje}{iu}
\icmlauthor{Serge Koudoro}{iu}
\icmlauthor{Nasim Anousheh}{iu}

\end{icmlauthorlist}

\icmlaffiliation{iu}{Department of Intelligent Systems Engineering, Indiana University Bloomington, USA}

\icmlcorrespondingauthor{Eleftherios Garyfallidis}{elef@indiana.edu}
% \icmlcorrespondingauthor{Eee Pppp}{ep@eden.co.uk}

\icmlkeywords{Machine Learning, ICML}

\vskip 0.3in
]



\printAffiliationsAndNotice{}  % leave blank if no need to mention equal contribution

\begin{abstract}

Clustering is a fundamental problem in machine learning where distance-based approaches have dominated the field for many decades. This set of problems is often tackled by partitioning the data into K clusters where the number of clusters is chosen apriori. While significant progress has been made on these lines over the years, it is well established that as the number of clusters or dimensions increase, current approaches dwell in local minima resulting in suboptimal solutions. In this work, we propose a new set of distance threshold methods called Theta-based Algorithms (ThetA). Via experimental comparisons and complexity analyses we show that our proposed approach outperforms existing approaches in: a) clustering accuracy and b) time complexity. Additionally, we show that for a large class of problems, learning the optimal threshold is straightforward in comparison to learning K. Moreover, we show how ThetA can infer the sparsity of datasets in higher dimensions.

%Distance-based clustering methods have been very popular in the past few years. Among the methods, k-means and its variety have been mostly used for Clustering, with good results in many datasets. Nevertheless, k-means based methods have their fundamental problem of local minimum, time complexity and need for a prior of number of clusters, which may not be available in many cases. In this paper, we describe how using distance thresholds instead of number of clusters is advantageous and provide an online distance threshold based method with a solution to the/ ordering problem. In most datsets our methods outperforms and is faster than traditional clustering methods.

%This document introduces Theta Grouping (TG). 

% This document provides a basic paper template and submission guidelines.
% Abstracts must be a single paragraph, ideally between 4--6 sentences long.
% Gross violations will trigger corrections at the camera-ready phase.
\end{abstract}

\section{Introduction}
\label{introduction}
Clustering is a an unsupervised learning technique used in a variety of domains such as collaborative filtering \cite{ungar1998clustering}, trend analysis \cite{aghabozorgi2015time}, multi-modal data analysis, social network analysis, biological data analysis, signal processing, etc. Given the wide range of applications, the problem of clustering has been tackled using different approaches such as dimensionality reduction, density estimation, probabilistic methods, spectral methods and distance-based techniques. Among all these methods, distance-based techniques (e.g. K-Means\cite{lloyd1982least}) have been used most widely on account of their simplicity and tractability. Distance-based clustering methods rely on providing apriori information to the algorithm. Based on this, one can further divide distance-based methods into three categories: 1) assuming number of clusters as known in advance, 2) a distance threshold as known or 3) by assuming a limiting number of data points  belonging to each particular cluster.

While clustering algorithms primarily focus on accurately partitioning the data, they also aimed at inferring information from a data exploration standpoint. In this work, we primarily focus on distance-based clustering given its broad adoption and propose a new framework, ThetA, which uses a distance threshold as an apriori parameter. The primary motivations of proposing ThetA is the improved accuracy, speed and robustness of the clustering obtained with simplification on parameter selection. We show this theoretically and experimentally by comparing ThetA against other commonly used clustering methods such as K-Means++ \cite{arthur2006k} and BIRCH \cite{zhang1996birch}. 

% We also show that ThetA works on any $\ell_p$ norm objective and compare the clusters using similarity metrics such as normalized mutual information.

Inference based on clustering is of cardinal importance as it is often applied as the first step in the machine learning pipeline. Information retrieved from clustering is often fed into subsequent learning algorithms. In many applications such as recommendation systems \cite{lu2015recommender}, biomedical analyses \cite{xu2010clustering}, anomaly detection \cite{agrawal2015survey}, etc. clustering is applied on a transient dataset, requiring the number of clusters to change. 

Our primary focus for ThetA stems from the fact that in scientific domains, we often have prior knowledge of the clusters we intend to find such as the physical dimensions of cells, atoms, bacteria, particles, etc. In such cases, it is much more useful to infer the number of clusters at multiple scales via a distance threshold rather than setting a fixed limit on the number of clusters in the data. In such datasets often the number of clusters is in the range of thousands which is hard to predict. Using a distance threshold had shown some success with different methods proposed in the past such as Leader Algorithm \cite{rush1988leader}, BIRCH \cite{zhang1996birch}, BSAS \cite{theodoridis_koutroumbas_2006}, QuickBundles \cite{garyfallidis2012quickbundles} and SL \cite{patra2011distance} but many challenges such as sampling ordering issues are still open.

The secondary motivation for proposing this framework derives inspiration from how humans scan images using saccades to form a sense of structure. By randomly sampling parts of an image, humans start clustering parts of it together. This has been studied in theoretical and systems neuroscience \cite{mcfarland2015saccadic}. 

\begin{figure}[ht]
\vskip 0.2in
\begin{center}
\centerline{\includegraphics[width=\columnwidth]{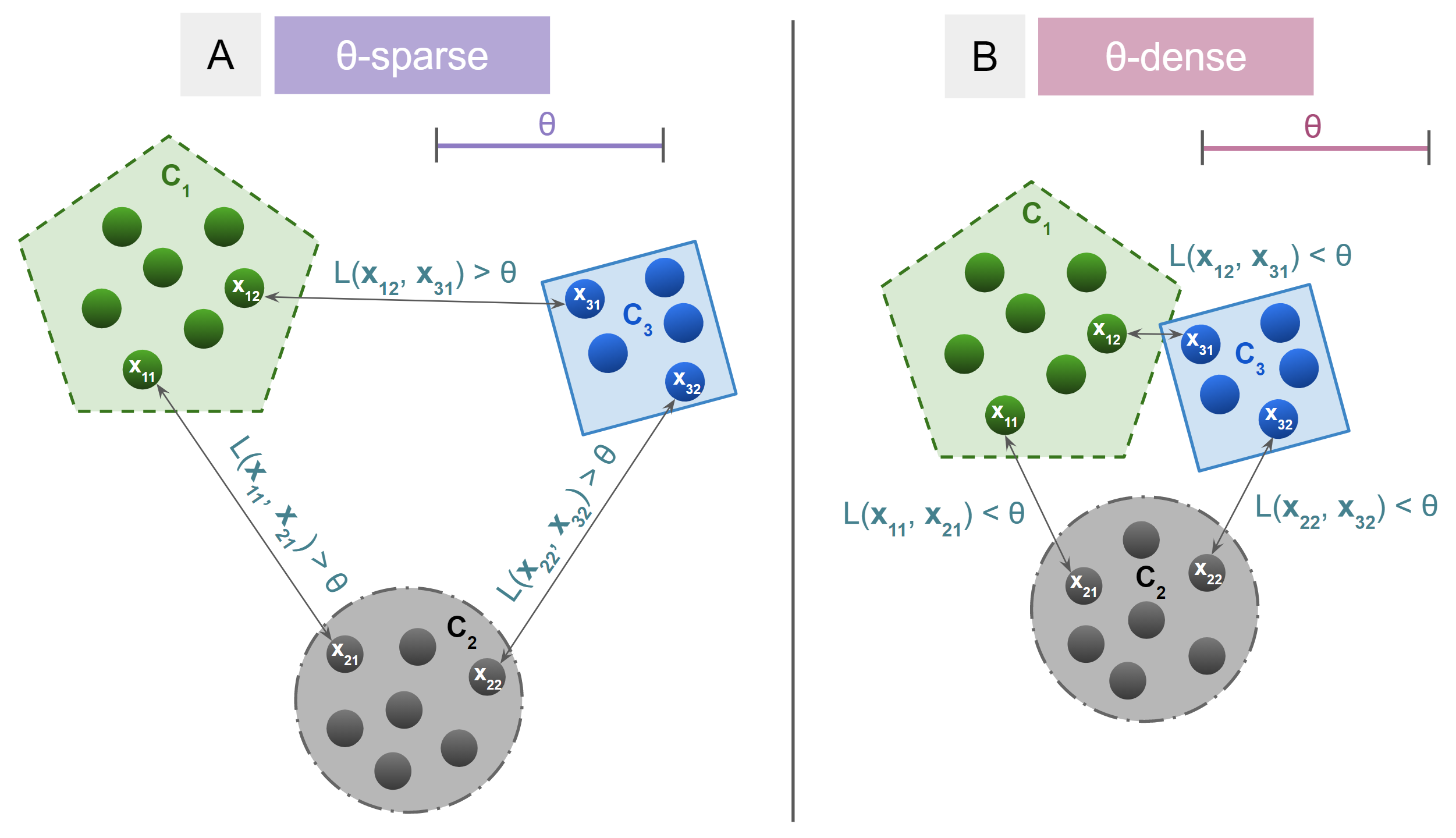}}
\label{fig:easy_hard_case}
\caption{Example diagram elucidating the difference between  $\theta$-sparse and $\theta$-dense clustering problems. If at least one pair of points from two different clusters have distance less than $\theta$ then we consider this as $\theta$-dense dataset otherwise a $\theta$-sparse. In the easier $\theta$-sparse case, our proposed approach will cluster the data in a single pass. In the harder $\theta$-dense case it will cluster in more passes at a low computational cost.}
\end{center}
\vskip -0.2in
\end{figure}

\section{Related Work}
\label{related_work}
\subsection{K-Partitional Clustering}
Distance based partitional clustering is most commonly addressed as a problem of grouping the data into K clusters, often using the means as the cluster representative, i.e. K-Means. These type of problems often require iterative optimization of an objective function until convergence. Since this set of problems is highly dependent on the initialization strategy, a variety of initialization methods have been proposed in the past \cite{hartigan1979ak}, \cite{milligan1985examination}, \cite{bradley1998refining}. From these, the most widely used implementation is the one of K-Means++ \cite{arthur2006k} with effective initialization for K-Means.  

\subsection{Prior threshold-based clustering}
Methods pertaining to threshold-based sequential clustering have been used to partition the data for many decades. As opposed to K-partitional clustering, the idea here is to sequentially partition the data on the basis of a distance threshold. In threshold-based clustering, one does not try to solve an optimization problem, but rather branches out into new clusters based on the sequence in which the data appears. While these algorithms are much faster in complexity, a reason they have not been adapted widely is due to an issue with the data ordering. Previous algorithms taking this approach have not taken this problem in to account or haven't managed to propose a solution to it. In this work we propose a new way to address the ordering problem and compare the results experimentally against other methods.

\subsection{Dimensionality and Distances}
It is well established that clustering becomes harder as the number of samples, the number of dimensions per sample and the number of underlying clusters increase. In this work, we shed light on this problem and report the performance of different distance based algorithms such as K-Means, K-Means++ and BIRCH. We show that the algorithms proposed in this work highlight superior performance and increased accuracy in multiple scenarios.

\section{Theory}
\label{theory}
\subsection{Notation}
We denote the number of samples $N\in{\mathbb Z^{+}}$, number of features (dimensions) $D\in{\mathbb Z^{+}}$, number of iterations $I\in{\mathbb Z^{+}}$, number of partitions $P\in{\mathbb Z^{+}}$. The data sets are denoted with $X$ and contain points $\textbf{x}\in\mathbb{R}^D$. The ThetA threshold is denoted with $\theta\in{\mathbb R_{\ge 0}}$. The number of clusters is denoted with $K\in{\mathbb Z^{+}}$. All distances that will be used are metric distances that satisfy the triangle inequality. Primarily we will use the $L^2$ norm (Euclidean distance) denoted simply as $L$.

\subsection{Fundamentals}

\textbf{Definitions} $\theta$-sparse: We define $\theta$-sparse as a clustering problem where the minimum inter cluster distance is $>\theta$. $\theta$-dense: We define $\theta$-dense as a clustering problem that is not $\theta$-sparse. %We define a sub-class of $\theta-dense$ as $\theta-super-dense$ when all minimum inter class distances are $<\theta$.

\textbf{Theorem 1}. Given a clustering problem with clusters $C_1, C_2, \dots, C_K$. If there are no pairs of samples $\textbf{x}_i, \textbf{x}_j$ from $C_i$ to $C_j$, $i\neq j$ where $\textbf{x}_i \in C_i$ and $\textbf{x}_j \in C_j$, that have a distance  $< \theta$, then Algorithm 1 will identify the exact clusters with a threshold parameter $\theta$ at exactly $N$ steps and the order of the selection of the samples will have no effect on the final outcome. 
%\end{theorem}

%\begin{proof}
\textbf{Proof}. In Algorithm 1 all the distances are computed either inside a cluster or between clusters. All the inside cluster distances will be $<\theta$ and all the distances between clusters will be always $>\theta$. Therefore, there are no cases were clusters are created between the actual clusters given that there are no pairs of samples $\textbf{x}_i, \textbf{x}_j$ from cluster $i$ to cluster $j$ that have a distance $<\theta$. \hfill $\blacksquare$
%\end{proof}

This brings us to an equivalence relationship between how the algorithm performs and how sparse or dense the data are. 

\textbf{Lemma 1} If Algorithm 1 is repeated using uniform random orderings of the samples and the labels do not change in any repetition then $X$ is $\theta$-sparse.

\textbf{Proof} By changing the ordering of selection we increase the probability to choose points across clusters that have distance $<\theta$ and therefore if this never happens $X$ is $\theta$-sparse. \hfill $\blacksquare$

We consider this case described in Theorem 1 as the easy case of the distance-based clustering problem. Notice that Theorem 1 does not depend on the shape or number of the clusters. 

The clustering problem becomes harder as the distances between points across the clusters get smaller (see Fig.~1). Meaning, that the order of the selection of the samples is more important than before. We present Algorithm 2 which takes that into account and provides a solution for the hard case. 

If at least one pair of samples from $\textbf{x}_i, \textbf{x}_j$ from cluster $i$ to cluster $j$ have a distance $<\theta$ then the number of orderings will be inversely proportional to the distance between the clusters.

When re-ordering is truly important? When the probability of selecting pairs with distance less than $\theta$ is high. This is happening in practice when the minimum distance between $C_i$, $C_j$ are $<\theta / 2$.

Algorithm 1 and 2 consider linearly separable clustering problems which  are the main focus of this work. However, as an application we present Algorithm 3 which allows to solve nonlinear clustering problems by combining Algorithms 1 and 2.

\begin{algorithm}[ht]
   \caption{- \textbf{ThetA Sparse Grouping (TSG)}}
   \label{alg:sg}

\begin{algorithmic}
   \STATE {\bfseries Input:} data $X$ of size $N$, threshold $\theta$.
   \STATE {\bfseries Output:} clustering $C$ of cardinality $K$ with centroids $\bar{C}_k$.
   \STATE $C_0$ $\leftarrow$ $X_0$, $K=1$, $\bar{C}_0=X_i$ \COMMENT{First point, first cluster}
   \FOR{$i=1$ {\bfseries to} $N-1$}
   \STATE x2c = inf($K$) 
   \FOR{$k=0$ {\bfseries to} $K-1$}
   \STATE d = dist($x_i, \bar{C}_{k}$) \COMMENT{Distance of point from centroid $k$}
   \IF{d $<=$ $\theta$}
   \STATE x2c[k] = d
   \ENDIF
   \ENDFOR
   \STATE $m$ = min(x2c); $a$ = argmin(x2c)
   \IF {$m <= \theta$}
   \STATE $C_a$ $\leftarrow$ $X_i$  \COMMENT{Insert data point to existing cluster} 
   \ELSE
   \STATE $K = K + 1$; $C_{K-1}$ $\leftarrow$ $X_i$ \COMMENT{Create new cluster}
   \ENDIF
   \ENDFOR
\end{algorithmic}
\end{algorithm}

\subsection{Complexity analysis}

%Algorithm 1 has a time complexity that depend on the number of samples N and the number of estimated clusters K. We assume here that most of the computation is taken for the calculation of distances between samples and centroids. The worst case complexity takes places when every data point belongs to a  different cluster (all singleton clusters). In such an event the worst time complexity can be calculated as O($(N^2 + N - 2)/2$). This is because the number of operations is increasing as a classical divergent series of this form $1+2+...+N$ minus the first sample for which distances are not calculated. The best time complexity is O($N$) when there is only one cluster. Given that in many problems $K\ll N$ we can consider this algorithm as a linear time. Apart from $N$ and $K$ the algorithms also depends on the size of the features $D$ which for now we consider as constant.

Algorithm 1 has a time complexity that depends on the number of samples N and the number of estimated clusters K. We assume here that most of the computation is taken for the calculation of distances between samples and centroids. The worst case complexity takes places when every data point belongs to a  different cluster (all singleton clusters). In such an event the worst time complexity can be calculated as O($(N^2 + N - 2)/2$). This is because the number of operations is increasing as a classical divergent series of this form $1+2+...+N$ minus the first sample for which distances are not calculated. The best time complexity is O($N$) when there is only one cluster. It is also simple to show that given $K$ predicted clusters, the worst time complexity in terms of ordering would be O($NK$). Given that in many problems $K\ll N$ we can consider this algorithm as a linear time. Apart from $N$ and $K$ the algorithms also depends on the size of the features $D$ which for now we consider as constant.

Assuming most memory is spent on saving centroids, Algorithm 1 has a best case space complexity of O($1$) and worst case of O($2N$) which takes place when all clusters are singleton clusters.

Algorithm 2 builds on top of Algorithm 1 but now the time complexity also depends on the number of iterations $I$. Therefore, worst case is now O($I (N^2 + N - 2)/2$) and best case is O($I N(N+1)/2$). The space complexity of algorithm is best case O($I$) and worst case O($2IN$).   

A reminder here that K-Means time complexity according to Lloyd's algorithm 2 is O($NKDI$), where $N$ number of points, $K$ number of clusters, $I$ number of K-Means iterations and $D$ number of dimensions. 

%Algorithm 1 (ThetA Sparse Grouping) calculates distance from each point to each centroid. One centroid is updated everytime, so it takes $O(N)$ time for updates where N is the number of data points. Given the optimal threshold $\theta$, $K^{\prime} \simeq K$ where K is the number of the intended clusters and $K^{\prime}$ is the number of clusters the algorithm predicts. Since the algorithm depends on the order of points, in comparison to Lloyd's algorithm we can think another worst case scenario when the first $K^{\prime}$ points chosen from data all serve as individual centroids. We would initially calculate distance from each point to 1 to $K^{\prime}$ centroids and continue to calculate distance from each point to $K^{\prime}$ centroids for roughly $N-K^{\prime}$ times. This gives the time complexity of $O(N K^{'} - \frac{K^{\prime 2}}{2} - \frac{K^{\prime}}{2}) + O(N) = O(N K^{'})$. Then it is trivial to show the time advantage to K-Means since it lacks the iteration term.

Algorithm 1 (ThetA Sparse Grouping) has clearly lower time complexity than Lloyd's algorithm. Similarly Algorithm 2 (ThetA Dense Grouping) will be slower than Lloyd's algorithm only if the iteration term is significantly larger than that of Lloyd's. Space complexity is also very low as the primary use of memory is to just store the centroids.

\section{Algorithms}
\subsection{ThetA Sparse Grouping (TSG)}
\label{theta_grouping}

ThetA Sparse Grouping (TSG) is shown in Algorithm 1. This is a sequential algorithm where clusters are created if the samples are far from existing clusters otherwise they are inserted to the closest already available  cluster. This algorithm is inspired by the Leader algorithm, BSAS, BIRCH and QuickBundles. We would like to emphasize that TSG processes each sample only once with a single parameter $\theta$.

%Very importantly the only parameter is $\theta$ and TSG processes each sample only once. 

\subsection{ThetA Dense Grouping}
\label{theta_merging}

ThetA Dense Grouping (TDG) is presented in Algorithm 2. TDG shuffles the data at each iteration, runs TSG and collects all centroids for each run. It then runs another TSG on all collected centroids and reassigns all points to the final centroids. The algorithm assumes that the centroids will be sparser than the original data which is a sensible assumption as TSG returns on average case less centroids $K$ than the original number of samples $N$.

%The number of iterations I depends on how much 

% Think of two clusters A and B. We make a few assumptions first, that can be proven trivial with more concrete proofs; cluster A and B have the same Gaussian distribution with a different mean, we have an optimal threshold $\theta$. Given we have an optimal threshold, we would not have more than one centroid within cluster. So the hard case would only be from where a point in cluster A is clustered with a point in B and vice versa. It is also important to note that there are 5 kind of centroid subgroups possible from the two clusters, where 3 subgroups come from a incorrectly clustered cases and 2 come from correctly clustered cases. @JS your stuff is commented here

For instance, say we had two clusters $C_A$ and $C_B$, assume they were randomly sampled from a Gaussian distribution with a different mean. Given an optimal threshold $\theta$, we would obtain one unique centroid for each cluster. For such a setting to be categorized as `dense' for ThetA, it would imply that a data point in $C_A$ was assigned to cluster $C_B$ and vice versa. In this type of a problem, each data point would fall in either one of the clustering cases, 2 - True positives and 3 False Positives, as discussed below. 

% We will divide this 'hard case' into two cases where a point in cluster A is not within threshold range to the correct centroid of cluster B and where it is within threshold.

To further elaborate on the problem, let's divide data points $\{\textbf{x}_1, \textbf{x}_2, \dots \textbf{x}_n\} \in X$ in the `dense' case into two sub-types: $C_{AB1}$ are the set of points whose pair-wise distances are less that the distance threshold $\theta$ and $C_{AB2}$ would be the set of points whose pair-wise distances are greater than $\theta$. 

% In the first case, let's assume among the number of shuffles, cluster A and B were correctly clustered $n$ times and incorrectly clustered $m$ times. Then originally, TSG would have a wrong cluster rate of $\frac{m}{n+m}$. However, if we look at the case where we try to cluster those centroids, a centroid in subgroup of centroids that merge part of cluster A and cluster B together, can be divided into two smaller subgroups where one would belong to cluster A and the other would belong to cluster B. 

\begin{figure}[ht]
\vskip 0.2in
\begin{center}
\centerline{\includegraphics[width=\columnwidth]{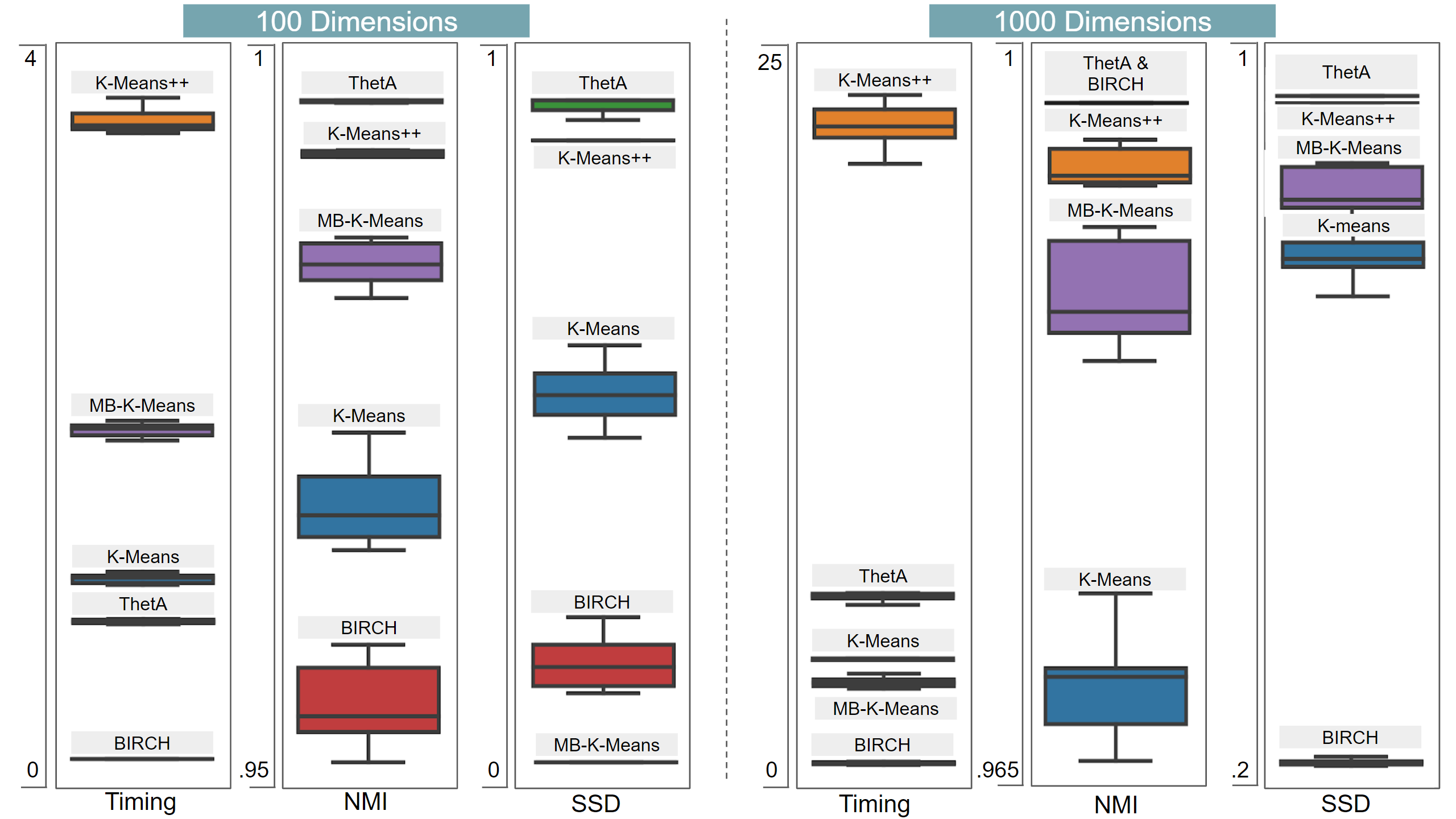}}
\label{fig:box_plot}
\caption{Depicts box-plots for comparing the performance of ThetA, K-Means, K-Means++, Minibatch K-Means (MB-K-Means) and BIRCH. The comparison has been performed on the basis of timing (seconds), Normalized Mutual Information (NMI) and Sum of Squared Differences (SSD) metrics. Higher values of NMI and SSD metric indicate better clustering quality and the lower value of timings indicates faster execution time.}
\end{center}
\vskip -0.2in
\end{figure}

In the case $C_{AB1}$, say the data points in $X$ were correctly assigned $n$ times and incorrectly assigned $m$ times, where $n+m = I$ and I is the number of shuffels as per TDG algorithm. If TSG algorithm was applied at this stage on the data, the false positive rate would be given by $\frac{m}{n+m}$. However in TDG, where we try to cluster these centroids in $C_{AB1}$, they would define a region of $X$ where the clusters $C_A$ and $C_B$ merge. This would imply that they could be assigned to either one of the two clusters.  
% If we consider the threshold $\theta$, it is trivial to show that from $n + 2m$ number of centroids that can be chosen, the wrong centroids to be paired with would be $m/2 + \epsilon$, where $\epsilon$ is a very small part from the opposite cluster's correct centroids. Then as long as $\frac{3m^{2}+mn}{m+n} > \epsilon$ holds, which can be easily proven since most of the correct centroids for the opposite cluster resides on the center of the cluster, the difficulty decreases on the final TSG step of TDG.

For a particular $\theta$, we can show that from $n + 2m$ number of centroids that can be chosen, the wrong centroids to be paired with would be $m/2 + \epsilon$, where $\epsilon$ is a very small part from the opposite cluster's correct centroids. Then as long as $\frac{3m^{2}+mn}{m+n} > \epsilon$ holds, which can be easily proven since most of the correct centroids for the opposite cluster resides on the center of the cluster, the difficulty decreases on the final TSG step of TDG.

Similarly in the case where $x_i \in C_{AB2}$, if we set the number of the true positive data points as $n$ and and false positives to $n - m$. With simple algebra, we can show that the clustering difficulty decreases as long as $m < \frac{4}{7}n$. Since this is the case when the center of the cluster is within the $\theta$ threshold, the condition will hold. This explains, why TDG works better on dense datasets than a single TSG.

%The error in the hard case we want to avoid is a case where a point in $C_A$ is predicted to be in the same cluster as a point in $C_B$. In that sense, we can divide the possible predicted centroids from the two clusters into five subgroups. When we find the wrong centroid, there would be 1. centroid from the remainder of cluster A, 2. centroid for the wrongly clustered points from A and B, 3. centroid from the remainder of cluster B. When we find the correct centroid, we have 4. centroid from cluster A and 5. centroid from cluster B. If we denote the case of wrong clustering as $m$ and correct clustering as $n$, subgroup 1 to 3 would have n points in each subgroup and subgroup 4 and 5 would have m. So the rate of wrong clustering is $\frac{n}{n+m}$. However, in the second ThetA Grouping step, the part of subgroup 3 that belong to cluster A should be clustered to points in 1, 2 and not the other part of 3. 4 and 5 would be unreachable or will have a very low probablity of being selected. The other part of subgroup 3, which belongs to B, should be clustered to points in 4, 5 and not the other part of 3. Note that subgroup 4, 5 in the first case and subgroup 1, 2 in the second case are unreachable or have a very low probability of being clustered with 3 due to the threshold. is not reachable each case due to the theta threshold. Using the assumptions, we can say that rate of predicting the wrong case will become $\frac{m}{n+2m}$, thus the problem has reduced difficulty by a rate of about $\frac{n+m}{n+2m}$.

\begin{algorithm}[ht]
   \caption{- \textbf{ThetA Dense Grouping (TDG)}}
   \label{alg:dg}

\begin{algorithmic}
   \STATE {\bfseries Input:} data $X$ of size $N$, threshold $\theta$, iterations $I$.
   \STATE {\bfseries Output:} clustering $C$ of cardinality $K$ with centroids $\bar{C}_k$
   \STATE $\bar{Z} = \emptyset$ \COMMENT{Set collects all centroids - Initially empty}
   \FOR{$iter$$=0$ {\bfseries to} $I-1$}
   \STATE shuffle $X$
   \STATE $\bar{Z}_{iter}$ $\leftarrow$ TSG($X$, $\theta$) \COMMENT{Generate centroids}
   \STATE append $\bar{Z}_{iter}$ to $\bar{Z}$ 
   \ENDFOR
   \STATE $Z$ $\leftarrow$ TSG($\bar{Z}$, $\theta$) \COMMENT {Cluster all centroids from previous step and generate new centroids}
   \STATE assign original $X_i$ to closest centroid of clustering $Z$
   \STATE update final centroids $\bar{C}_k$ of final clustering $C$
\end{algorithmic}
\end{algorithm}

\subsection{ThetA Nonlinear Chaining (TNC)}
\label{theta_chaining}
Algorithm 3 is an application example of how we can use TDG for nonlinear clustering problems. The idea is that first we start with TDG to produce small clusters and then we chain clusters together that are close to each other. A simulated example with comparisons is provided in Supplementary Material using the publicly available Shape Sets dataset. Note that centroids for clusters are not updated (new centroids are not created) as most clustering algorithms do with iterations. Instead, we keep all the centroids that are chained together to belong to that cluster and return new larger clusters at the end.
One iteration example of TNC holds of chaining\_list of $[[0],[1],[2],[3],[4],[5]]$ elements with $K = 6$ and centroid indices $[0-5]$. Updated chaining\_list will join some of the clusters $[[0,4],[1,2],[2,1],[3],[4,0,5,0],
[5,4,0]]$. After that relabeling phase starts that keeps only the smallest indices in the chaining\_list such that $[[0],[1],[1],[3],[0],[0]]$ \\
Finally, the updated centroid indices are $[0,1,1,3,0,0]$ and now the actual number of clusters is $K' = 3$ as many clusters have been merged together.
\begin{algorithm}[ht]
   \caption{- \textbf{ThetA Nonlinear Chaining (TNC)}}
   \label{alg:cn}

\begin{algorithmic}
   \STATE {\bfseries Input:} data $X$ of size $N$, threshold $\theta$, neighborhood $\epsilon$, iterations $I$ and $I_2$.
   \STATE {\bfseries Output:} clustering $C'$ of cardinality $K'$ \STATE $C$ $\leftarrow$ TDG(X, $\theta$, $I$) \REPEAT
   \STATE Initialize chaining\_list (size $K$) 
   \STATE \COMMENT{Above contains one set per id of centroid}
   \IF {dist between two centroids $<$ $\epsilon$}
   \STATE add both centroids ids to chaining\_list
   \ENDIF
   \STATE merge and update chaining\_list \COMMENT{transitively}
   \UNTIL total number of new clusters does not change
   \STATE assemble new clustering $C'$ using chaining\_list
\end{algorithmic}
\end{algorithm}

\section{Experiments and Comparisons}
In order to compare the performance of ThetA with other clustering algorithms, we setup a simple experiment with 2D point clouds randomly sampled from a normal distribution. We further divide the experiment into $\theta$-sparse and $\theta$-dense cases as described in Sec.~5.1 and 5.2. We compare ThetA with commonly used clustering algorithms such as K-Means, K-Means++, BIRCH and Mini-batch K-Means \cite{sculley2010web}. We compare the clustering accuracy against the ground truth of the simulation using the normalized mutual information (NMI). We also propose using the sum of squared differences (SSD) metrics which evaluates the correctness of centroids. We use SSD to report the percentage of correctly identified ground truth centroids up to numerical precision. Lastly, we also compare the actual running time for each of the algorithms on a standard i7 CPU with 16GB RAM. Scikit-Learn \cite{scikit-learn} package (v. 0.20.4) was used for methods other than ThetA.
% Design experiments
\subsection{Easy Case}
\label{easy_case}
The experiment compares methods on easy case data, where we have 100 clusters and each centroid has a wide enough distance to be considered a $\theta$-sparse case as discussed above. The Euclidean distance ($L$) between centroids is 10. K-Means, K-Means++ and Mini-batch K-Means were all run with the number of clusters set to 100. BIRCH was run with threshold of 2.7 and branching factor of 50. TSG was run with threshold of 6.0. Each method does an accurate job on the clustering problem but ThetA has a considerable timing advantage to other methods. Detailed result and figure can be seen in the supplementary section.
\begin{figure*}[ht]
\vskip 0.2in
\begin{center}
\centerline{\includegraphics[width=1.0\textwidth]{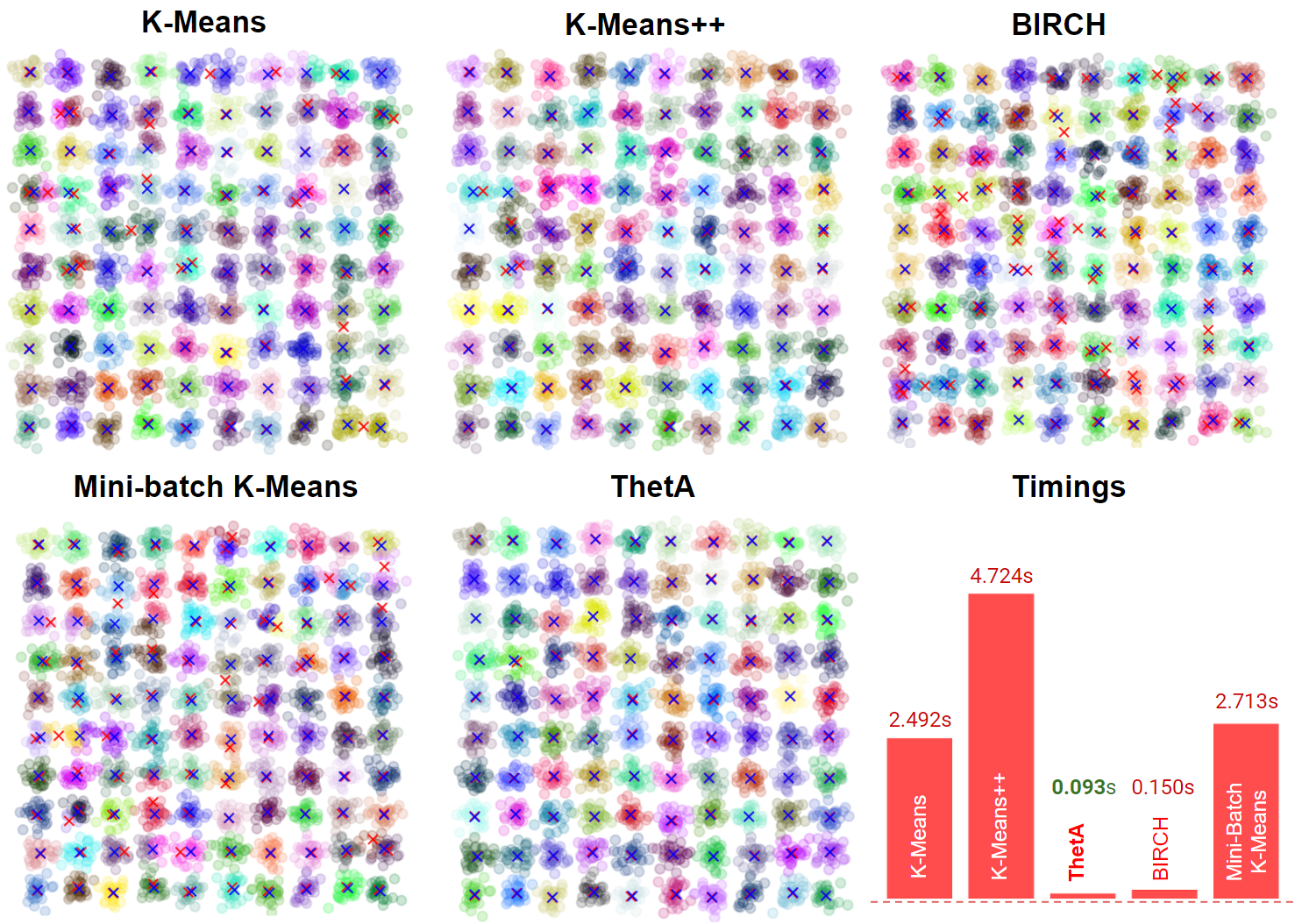}}
\label{fig:hard_case_100blobs}
\caption{Depicts the performance of ThetA against other algorithms for a hard case of clustering where the data-points are randomly sampled from a normal distribution. Blue crosses are the ground truth centroids and red crosses are the predicted centroids. The more red crosses are visible the worse the clustering prediction will be. Note that ThetA identifies all centroids correctly in the least amount of time.}
\end{center}
\vskip -0.2in
\end{figure*}
\begin{figure*}[ht]
\vskip 0.2in
\begin{center}
\centerline{\includegraphics[width=1.0\textwidth]{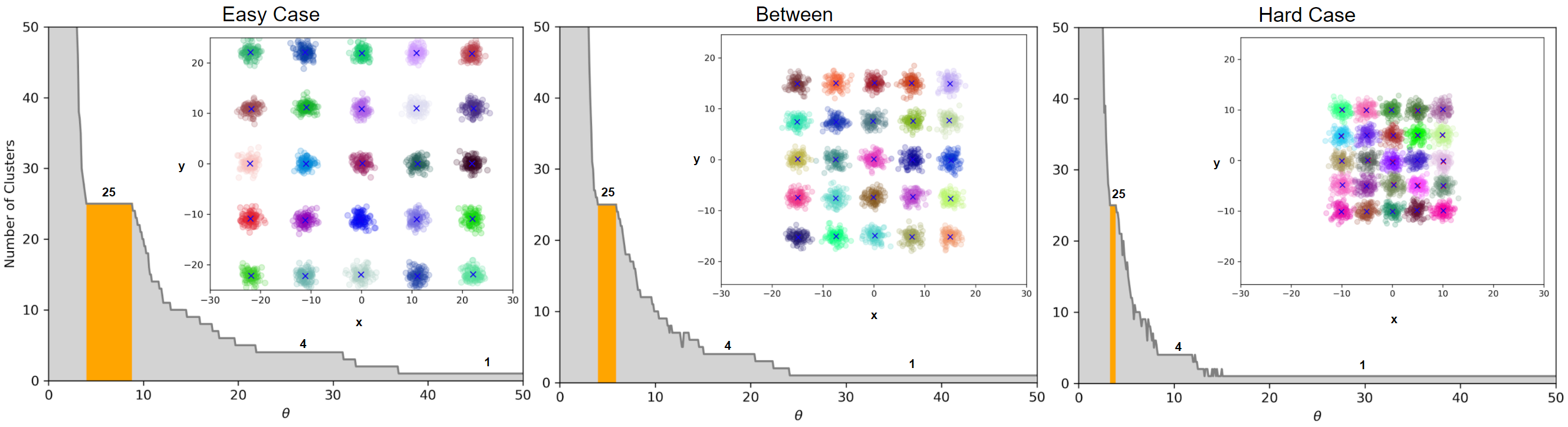}}
\label{thetarange}
\caption{Learning $\theta$ is straightforward as optimal ranges appear at roughly the same locations (see dark orange areas under the curve). Notice that optimal ranges are larger the sparser the problem is. Which in this case means the further the clusters are. The optimal number of clusters here is 25 which appears in at the upper size of the elbow. As the optimal range is reduced the harder the problem becomes. This is an indication that shuffling number $I$ of TDG should be increased as sampling ordering becomes more critical or that threshold $\theta$ should be reduced.}
\end{center}
\vskip -0.2in
\end{figure*}

\subsection{Hard Case}
\label{hard_case}
In the hard case experiment we have 100 clusters and each centroid has a short euclidean distance of 5 to each other. Unlike the easy case experiment, K-Means, K-Means++ and Mini-Batch K-Means were assigned 50 number of initializations additional to the number of clusters to match the advantage ThetA (TDG) would have by its multiple number of shuffles. BIRCH was run with threshold of 2.0 and branching factor of 50. ThetA used a threshold of 3.6 and 50 number of shuffles. The resulting Fig.~3 shows that now that the clusters are closer to each other, methods start to fail, while ThetA maintains its accuracy and speed advantage. It can also be seen in the supplementary section ThetA has a stable accuracy regardless of the difference in number of points per cluster, unlike other methods.

\subsection{Change in number of dimensions}
We provide box plots, as shown in Fig.~2 drawn from 10 runs of each case; 100 and 1000 number of dimensions with a fixed number of clusters set to 100. The experiments were done on the hard case, where the Manhattan distance difference between each consecutive cluster was set to $1.6D$ where D is the number of dimensions, for consistency. The number of shuffles for TDG and number of initializations for K-Means, K-Means++ and Mini-batch K-Means were all set to 50. On 100 dimensions the conditions were threshold of 15.4 for ThetA and 12.75 for BIRCH and for 1000 dimension the thresholds were set to 50 and 29.96 for ThetA and BIRCH. The experiment with the change in number of clusters has been added to the supplement.

With higher number of dimensions, while the time taken for ThetA can be slower than some methods in higher dimensions (which is possibly because these algorithms converge faster to a local minimum solution rather to the global solution), we can see that the accuracy of TDG remains stable while other methods fluctuate or decrease.

\subsection{Choice of Theta}
In Fig.~4 we show an experiment to elucidate how the optimal value of theta change in relation to the distances between the clusters. We basically generated normal distributions on a grid and simply repeated TDG for different distance thresholds. Notice that we have a wide range of optimal $\theta$ that provide the correct number of cluster. Notice also the changes in optimal ranges of optimal $\theta$ given the sparsity of the dataset. As density increases, the range of optimal $\theta$ decrease, but there is still a relatively wide range of $\theta$. 

For the easy case it becomes trivial to learn the optimal $\theta$ and ordering is not a consideration but as the problem becomes harder the optimal $\theta$ becomes harder to find and the ordering problem becomes more severe. However, even in the case of a super dense problem we can use a lower $\theta$ value and study the sizes of the clusters. See example and comparisons using S1-4 datasets in Fig.~6. The distances between centroids were 11, 7.5 and 5 for each case.

\subsection{Deep Embedded Clustering}
\label{deep_embed_exp}
We show how Algorithm 2 can facilitate and boost Deep Embedded Clustering \cite{xie2016unsupervised} in a completely unsupervised way. We follow the paper's base model architecture from training a simple Autoencoder model with 4 dense layers for the encoder and 4 dense layers for the decoder. For this application we use real data from the well studied MNIST dataset \cite{lecun1998gradient}. The method adds a clustering layer which uses the last layer of the encoder as a latent feature dimension input and produces soft labels using the Student t-distribution. KL divergence and Mean Squared Error loss are used for reconstruction and clustering loss. For comparison, we initialize the centroids in the clustering layer using three methods; K-Means, K-Means++ and TDG with threhold 4.724. Since there are many outliers that can be detected, we cluster more than 10 centroids using Algorithm 2 and use only the top 10 centroids that have the most samples in their cluster. As seen in Fig.~5, ThetA starts with a higher accuracy and NMI but also has a higher rate of increase through epochs. Note that this is not an experiment to show how the TDG performs using latent dimensions but to compare each method's ability to cluster latent features.

\subsection{Set-S dataset}
\label{set_s_exp}
We run TDG and K-Means++ on the Set-S dataset \cite{Ssets}. The dataset contains clusters of different shapes. We show how theta can correctly assign the clusters given an optimal threshold. Moreover, it also helps in inferring the number of clusters in each of S1-4 shown in Fig.~6. As seen in the line plot, there is a notable drop in the points per cluster after passing the correct number of clusters. We further elaborate on this observation in the case of Deep Embedded Clustering experiment where we can identify the correct number of clusters by using a threshold that creates more than the number of correct clusters. NMI table shows that the resulting clusters are closer to the ground truth than K-Means++. K-Means++ setup had 10 number of initializations with 300 max iterations. TDG was used with $\theta$ equal to $1e5$ for all S1 to S4 datasets.

\begin{figure}[ht]
\vskip 0.2in
\begin{center}
\centerline{\includegraphics[width=\columnwidth]{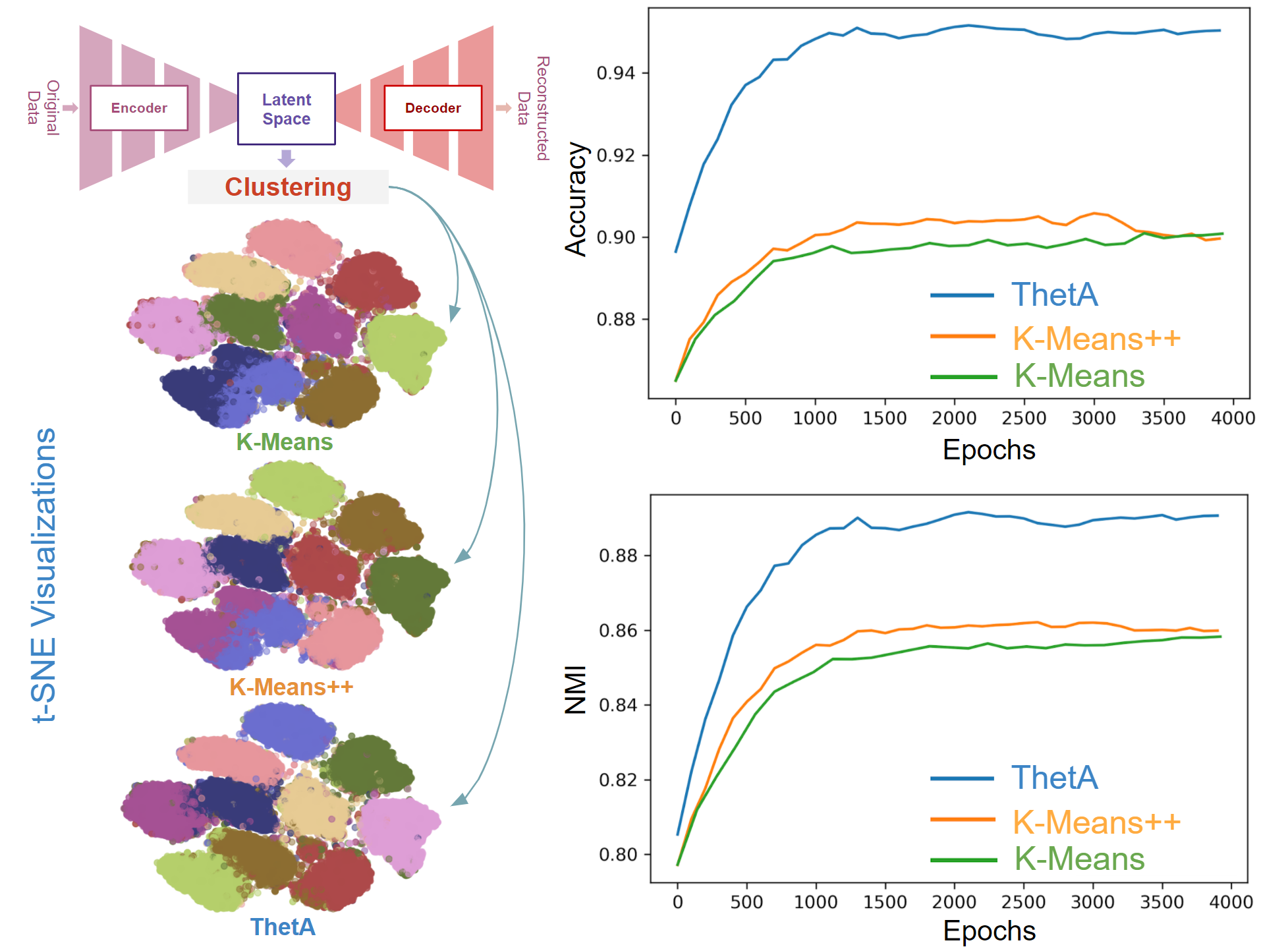}}
\label{fig:deep_embed}
\caption{Depicts a comparison of ThetA with K-Means and K-Means++ for Deep Embedded Clustering. The t-SNE plots of the clustering have been presented for each of the methods for qualitative comparison of the results. Each method has been evaluated using the NMI metric and accuracy.}
\end{center}
\vskip -0.2in
\end{figure}

\section{Results Summary}
Our results suggest that ThetA (TDG \& TSG) tends to surpass other methods in terms of accuracy and complexity in both linearly separable easy and hard cases, regardless of the numbers of dimensions. Also, though not always faster than K-Means, the algorithm always takes less time than K-Means++, the only comparable method in the experiments that has a relatively close accuracy. Note that we compare against the Elkan's method \cite{elkan2003using} for both K-Means and K-Means++ which is significantly faster than Lloyd's algorithm. Also note that BIRCH is faster in convergence with respect to ThetA, however, it does not converge to the correct solution.

As seen in the choice of ThetA experiment, though the range of stable $\theta$ decrease as the problem gets harder, it still shows that there is a large enough region that predicts the correct number of clusters which makes it easier to learn the correct $\theta$ for ThetA algorithms to use.

It is also important to note that finding more than ground truth clusters through TDG can be easily solved by choosing only the centroids with the highest number of points in their clusters. The sharp drop in Fig.~6 suggests that this number and its corresponding $\theta$ is straightforward to learn.

\section{Discussion}
\begin{figure}[ht]
\vskip 0.2in
\begin{center}
\centerline{\includegraphics[width=\columnwidth]{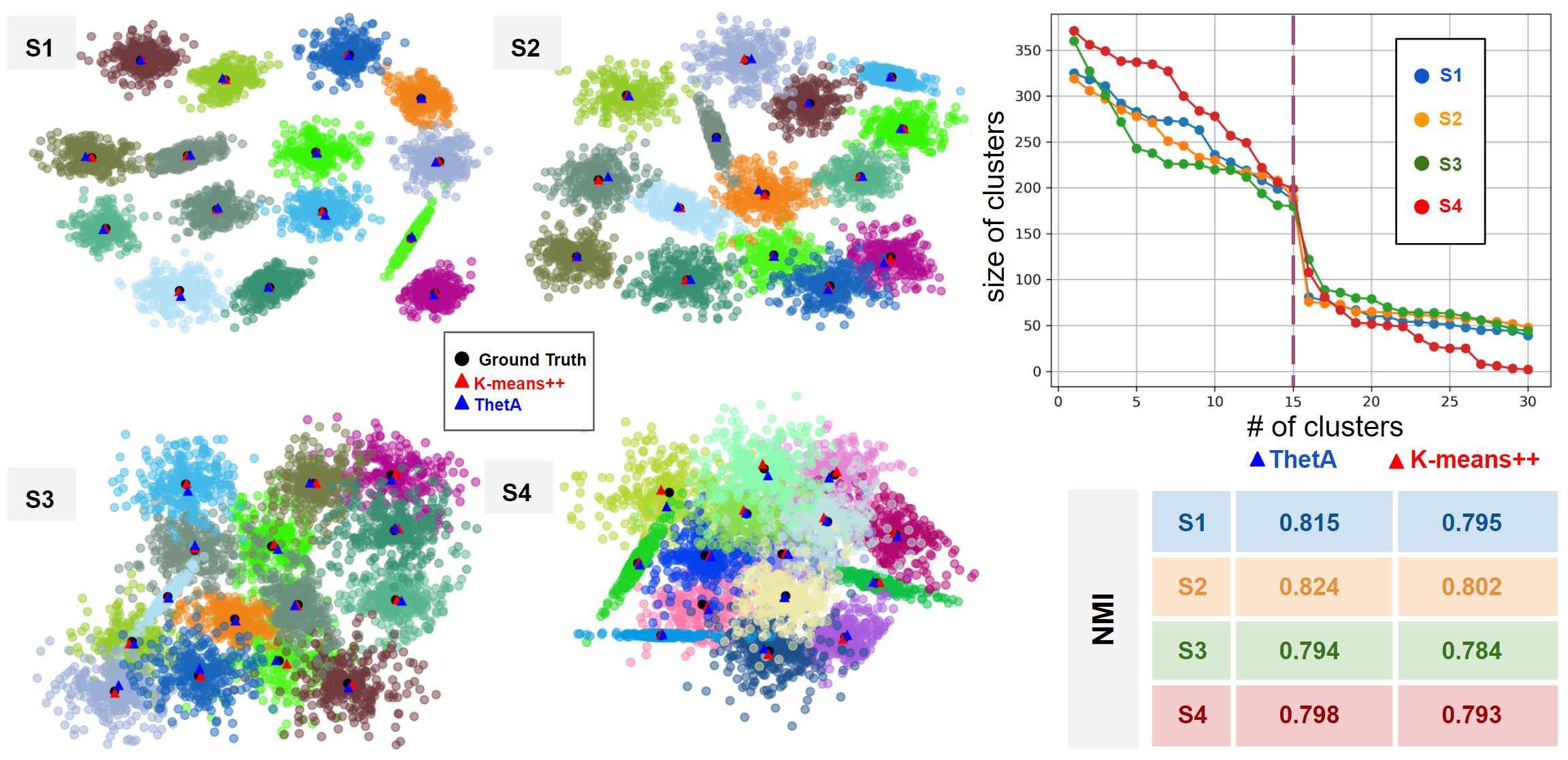}}
\label{fig:set_s}
\caption{We show how ThetA can be used to detect the number of clusters using the Set-S dataset and compare it against K-Means++. For each of S1 to S4 datasets we show how the correct number of clusters can be detected using an elbow plot. We also compare the two clustering methods using the NMI metric.}
\end{center}
\vskip -0.2in
\end{figure}

%Fast and robust unsupervised method as shown by our experiments. ThetA performance can increase by ideas proposed by Elkan, Hamerly, XX or parallelization. 

ThetA is a fast and robust unsupervised method as shown in the complexity analysis and experiments. This is an introductory paper to introduce new distance threshold based algorithms and propose using distance threshold rather than number of clusters for solving linear and in the future non-linear problems. There are still many ideas that can further improve ThetA, including workarounds to decrease the number of distance calculations. Techniques used by Elkan K-Means, Hamerly K-Means \cite{hamerly2010making} or parallelization can be all used to improve ThetA's already fast execution time.

It is important to note that ThetA is very stable in terms of scaling. Sec.~5.3 shows this by increasing the number of dimensions and increasing the number of clusters. Additionally, our experiments show that the algorithms are less susceptible to datasets which have different number of points per cluster. These stable features of the algorithm reinforces our claim that ThetA should be used on biological and physical sciences data instead of other methods, not just because it uses only a distance threshold but also because of its stability. Experiments validating the claims can be seen in the main document and the supplementary sections.

%Introduces equivalence functions for identifying the sparsity/density of datasets in high and low dimensions.

%If multiple iterations of TSG do not change the outcome then this is certainly a $\theta$-sparse dataset and single run of TSG is an optimal identifier of the clusters. 

In addition, TSG is equivalent to identifying the sparsity/density of datasets in high and low dimensions. If multiple iterations of TSG do not change the outcome then the dataset can be considered sparse. Otherwise, it would be a dense dataset. 

There can be an ordering issue where TDG does not converge after multiple runs, which implies that correct clustering does not have a higher chosen rate. However, possibly a good solution can be generated by deploying a lower $\theta$ value and studying the cluster sizes as shown in Sec.~5.5 and 5.6.

%Applications for deep embedding clustering were shown providing significant improvement. Remarkable 94\% accuracy from a completely unsupervised experiment.

The experiment on the MNIST Dataset using Deep Embedded Clustering with TDG has a remarkable accuracy of 94$\%$, considering the approach is completely unsupervised. Since other deep clustering approaches \cite{caron2018deep, yang2017towards} also use K-Means as their clustering method, it is safe to assume that using ThetA will increase their performance likewise.

 By proposing TDG and TNC, we for the first time show how the ordering problem persistent in such distance-threshold based algorithms can be tackled. We provide a short proof-sketch on Sec.~4.2 of how TDG improves the clustering accuracy by accounting for this issue. With a surge of interest in applying clustering to deep learning problems, we show how ThetA can be used to perform clustering in the latent space via Deep Embedded Clustering in Sec.~5.5 with an improved accuracy and cluster assignment. While all the experiments in this work were performed using the $L^2$ norm, future extension of this work would show that ThetA can be extended for other $L^p$ norms.

\section{Conclusions}

The problem of identifying the structure of high dimensional spaces is one that has occupied scientists for decades. In this work, we introduce a new family of algorithms called ThetA which enable the study of these high dimensional spaces and their sparsity in a principled way. In addition, we stress that ThetA outperforms the state of the art methods in linearly separable clustering problems both in speed and accuracy. We also provide guidelines of how these algorithms can be used for solving nonlinearly separable clustering problems and boosting the accuracy of unsupervised deep learning solutions.

\bibliography{example_paper}
\bibliographystyle{icml2021}

% \appendix
% \section{More Figs++}

%%%%%%%%%%%%%%%%%%%%%%%%%%%%%%%%%%%%%%%%%%%%%%%%%%%%%%%%%%%%%%%%%%%%%%%%%%%%%%%
%%%%%%%%%%%%%%%%%%%%%%%%%%%%%%%%%%%%%%%%%%%%%%%%%%%%%%%%%%%%%%%%%%%%%%%%%%%%%%%

\end{document}